\pdfoutput=1
\documentclass[11pt]{article}

\usepackage{acl}

\usepackage{times}
\usepackage{latexsym}
\usepackage[T1]{fontenc}
\usepackage[utf8]{inputenc}

\usepackage{microtype}
\usepackage{amsmath}
\usepackage{amsfonts}
\usepackage{amssymb}
\usepackage{bbm}
\usepackage{hyperref}
\usepackage{nicefrac}
\usepackage{url}
\usepackage[capitalise]{cleveref}
\usepackage{xcolor}
\usepackage{amssymb}
\usepackage{graphicx}
\usepackage{wrapfig}
\usepackage{multirow}
\usepackage{multicol}
\usepackage{booktabs}
\usepackage{makecell}
\usepackage{subcaption}
\usepackage{algorithm}
\usepackage{algpseudocode}
\usepackage{enumitem}
\setitemize{noitemsep,topsep=2pt,parsep=3pt,leftmargin=10pt}

\title{\model: Dynamic Contextual Compression for Decoder-only LMs}

\author{
Guanghui Qin$^{\color{blue}\eta}$\thanks{\hspace{4pt}Work done in part during Guanghui Qin's internship at Microsoft Research.} \hspace{50pt}  Corby Rosset$^{\color{blue}\mu}$ \hspace{50pt} Ethan C. Chau$^{\color{blue}\mu}$ \\ \textbf{Nikhil Rao}$^{\color{blue}\mu}$ \hspace{50pt} \textbf{Benjamin Van Durme}$^{\color{blue}\eta,\mu}$ 
\vspace{10pt}\\
$^{\color{blue}\eta}$Johns Hopkins University \quad $^{\color{blue}\mu}$Microsoft \\
\texttt{\{gqin2,vandurme\}@jhu.edu} 
}

\newcommand{\llama}{\textsc{LLaMA}}
\renewcommand{\vec}{\mathbf}
\newcommand{\model}{\textsc{Dodo}\,}
\newcommand{\nugget}{\textsc{Nugget}\,}
\newcommand{\repr}{\texttt{nuggets}\,}
\newcommand{\compressive}{\textsc{Compressive}\,}

\newcommand{\full}{\textsc{Full}\,}
\newcommand{\summ}{\textsc{LMSumm}\,}
\newcommand{\nodoc}{\textsc{NoDoc}\,}
\newcommand{\thick}{\specialrule{.1em}{.05em}{.05em}}

\newcommand{\trans}{\mathtt{Transformer}}
\newcommand{\lmhead}{\mathtt{LMHead}}
\newcommand{\attn}{\mathtt{Attn}}
\newcommand{\topk}{\mathtt{TopK}}
\newcommand{\sco}{\mathtt{Scorer}}
\newcommand{\nugop}{\mathtt{Dodo}}
\newcommand{\stopg}{\mathtt{StopGrad}}

\newcommand{\istar}{$\mathcal{I}^*$\,}
\newcommand{\ibar}{$\bar{\mathcal{I}}$\,}

\begin{document}
\maketitle

\begin{abstract}
Transformer-based language models (LMs) are inefficient in long contexts.
We propose \model, a solution for context compression.
Instead of one vector per token in a standard transformer model, \model represents text with \emph{a dynamic number} of hidden states at each layer, reducing the cost of self-attention to a fraction of typical time and space.
Moreover, off-the-shelf models such as \textsc{LLaMA} can be adapted to \model by efficient parameter tuning methods such as LoRA.
In use, \model can act as either an autoregressive LM or a context compressor for downstream tasks.
We demonstrate through experiments in language modeling, question answering, and summarization that \model retains capabilities in these tasks, 
while drastically reducing the overhead during decoding.
For example, in the autoencoding task, \model shrinks context at a 20x compression ratio with a BLEU score of 98\% for reconstruction, achieving nearly lossless encoding. 
\end{abstract}

\section{Introduction}

Transformer-based LMs~\citep{transformers17} suffer from quadratic computational complexity w.r.t. sequence length, making it challenging to scale to long sequences.
Proposed solutions~\citep{efficient22} include sparsifying attention patterns~\citep{longformer20,ding2023LongNetScalingTransformers} or approximating the attention computation with kernel methods~\citep{performer21}. %
However, not all these approaches are proven effective for NLP tasks \citep{nlpeffective23}, and very few of them are applied to large language models (LLMs), such as LLaMA \citep{llama23}.

\begin{figure}
\centering
\includegraphics[width=7.5cm]{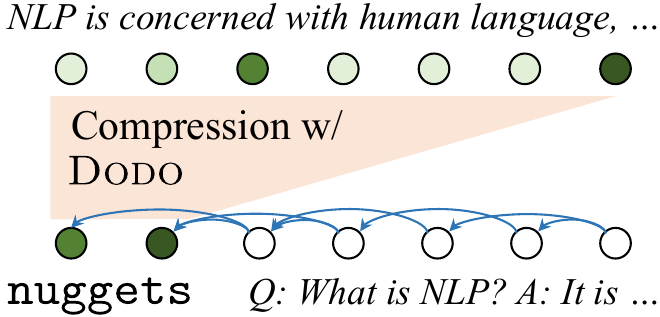}
\caption{
\model efficiently maps long inputs into a compressed set of vectors named \repr, which can then be attended to when processing a query. 
}

\label{fig:intro_fig}
\end{figure}

We propose \model, a solution for \underline{d}ynamic c\underline{o}ntextual compression for \underline{d}ecoder-\underline{o}nly LMs.
While a standard transformer  represents a text with vector sequences of the same length as tokens, the
intuition of \model is to use \emph{a smaller, variable number} of vectors as a contextual representation.
Past research indicates that a subset of token embeddings, named \repr, in an encoder with global attention may carry enough information to reconstruct surrounding context~\citep{nugget23}, and upon inspection those authors observed these \repr tended to account for \emph{preceding} text.
This suggests a decoder-only model might be dynamically capable of deriving such a representation online (\cref{fig:intro_fig}). 
To enable \model requires addressing a selection process that is not differentiable: we adopt the straight-through estimator~\citep{straightthrough13} to make the model end-to-end trainable.

Past work on context compression, such as \citet{icae24} and \citet{gist23}, appends \emph{fixed} \emph{additional tokens}.
\model \,\emph{grows} the representation with sequence length and \emph{re-uses} existing token embeddings.
Moreover, unlike pattern-based methods that evenly chunk the text~\citep{compressive20}, experiments show that \model spontaneously learns to use \emph{textual delimiters} as \repr, naturally splitting the text into subsentential units (\cref{sec:intrinsic}).

\model supports causal masking and can be naturally used as an autoregressive LM.
We experimentally demonstrate that \model can achieve a perplexity score lower than the original LM with restricted memory, outperforming the baseline model of \citet{compressive20}.
For tasks with a fixed context, e.g. long-form QA, \model works as a context compressor:
It encodes a token sequence into a shorter vector sequence, achieving a configurable compression ratio.
In experiments on autoencoding, we demonstrate that \model can achieve near lossless encoding with a compression ratio as high as 20x, a marked improvement over ICAE \citep{icae24}.
After fine-tuning,
\model is effective in downstream NLP tasks such as question answering (QA) and summarization,
where it performs on par with or even better than the original LMs while achieving a compression ratio as high as 10x.

In summary, we propose \model for contextual compression for decoder-only transformers.
It learns to subselect a fractional number of tokens as context representation.
A straight-through estimator ensures that \model is differentiable and can be trained with the next-token prediction objective.
\model achieves a remarkable compression ratio of up to 20x and is shown to be effective in tasks such as autoencoding, language modeling, and applications including QA and summarization.

\section{Approach}

In this paper, we study the language modeling problem $p(w_t\mid w_{<t})$, where $w_i\in V$ is a sequence of tokens and $V$ is the vocabulary.
The common Transformer~\citep{transformers17} approach encodes a token sequence $w_{1:n}$ into a sequence of vectors and then predicts the next token:
\begin{align}
\left(\vec{x}_1^{L}, \vec x_2^L \dots, \vec{x}_n^{L} \right)&= \trans_\theta(w_{1:n}),
\label{eq:transformer}\\
p(w_{n+1}\mid w_{1:n}) &\sim \lmhead_\theta(\vec x_n^L) \label{eq:lmhead},
\end{align}
where $\theta$ is the parameter, $L$ is the number of transformer layers, $\vec{x}_t^L\in\mathbb{R}^d$ is the hidden state of the $t$-th token in the $L$-th layer, $d$ is the hidden state dimension, and $\lmhead$ is a feedforward neural network that defines a categorical distribution over the vocabulary.
In the decoder-only transformers, $\vec x_t^{l+1}$ is encoded by attending to past token representation in the $l$-th layer:
\begin{align}
    \vec x_t^{l+1} = \attn_\theta(\vec{x}_t^{l},\vec{x}_{1:t}^{l}),~l = 1,2,\dots,\mathbin{{L}{-}{1}}
    \label{eq:originalattn}
\end{align}
where the $\attn$ function takes query and key (value) vectors as arguments.
\cref{eq:originalattn} can be inefficient with long sequences as its computation grows quadratically with the sequence length.
In this paper, we aim to answer: \emph{Can we find an alternative method to efficiently approximate $\vec{x}_t^l$ ?}

\subsection{Representing texts with \model}
\label{sec:n2dmethod}

In \cref{eq:originalattn}, context information up to the $t$-th token is encoded into $t$ vectors as hidden states.
Intuitively, we can reduce the computational overhead by controlling the size of hidden states.
Formally, we want to encode $t$ tokens $w_{1:t}$ into $k$ vectors: $(\vec z_1^{l}, \dots, \vec z_k^{l})$, where $k\le t$.
Following prior work \cite{nugget23} we refer to these vectors as \repr.
Then $\vec x_t^{l+1}$ is derived by
\begin{align}
    \vec x_t^{l+1} = \attn_\theta(\vec{x}_t^{l},\vec{z}_{1:k}^{l}),~ l = 1,2,\dots,\mathbin{{L}{-}{1}}.
    \label{eq:zattn}
\end{align}
Please note that \emph{$k$ is not a fixed number}~\citep{mvdoc22,icae24} \emph{but a dynamic number that depends on the input sequence $w_{1:t}$}. 
We will discuss the choice of $k$ later.

We observe that $\vec{x}_{1:t}^l$ encodes the information of tokens $w_{1:t}$,
thus one may derive $\vec z^l_{1:k}$ from $\vec{x}_{1:t}^l$.
We therefore select $\vec z_{1:k}^l$ by \emph{subselecting vectors} from $\vec{x}_{1:t}^l$.
Formally, we have (c.f. \S3.3 in \citealp{vcc23} and \S3.1 in \citealp{nugget23}):
\begin{align}
    \{\vec z_1^l, \dots, \vec z_k^l\} &= \{\vec x_i^l \mid \alpha_i = 1, 1\le i \le t \}, 
    \label{eq:zselection}
    \\
    p(\alpha_i = 1) &= \sigma (\sco_\varphi(\vec x_i^\iota))
    \label{eq:scorer},
\end{align}
where $\alpha_i$ is a binary variable indicating if $\vec{x}_i^l$ is selected, $p(\alpha_i = 1)$ refers to a Bernoulli distribution, $\sco_\varphi$ is a feedforward neural network parameterized by $\varphi$, and $\sigma$ is the sigmoid function.
$\sco_\varphi$ takes as input $\vec{x}^\iota_i$, the hidden state of $w_i$ in the $\iota$-th layer, where $\iota$ is a hyperparameter.
\footnote{We empirically set $\iota=3$ in all experiments.}
That is, tokens that were assigned with higher scores by $\sco$ is more likely be selected as \repr.

Note that $\iota$ in \cref{eq:scorer} does not depend on $l$, thus it selects the same set of indices for all the layers.
In the remainder of this paper, we abstract the process of 
\cref{eq:transformer,eq:zattn,eq:zselection,eq:scorer}
into a $\nugop$ operator:
\begin{align}
    \vec z_{1:k}^{1:L} = \nugop_{\theta,\varphi}(w_{1:t}),\quad 1\le k \le t. \label{eq:nugop}
\end{align}
We may omit the superscript and use $\vec z_i$ ($\vec x_i$) to indicate $\vec z^{1:L}_i$ ($\vec x^{1:L}_i$), the $i$-th \repr in all layers.

So far, we only assume that $k$ is a dynamic number depending on $w_{1:t}$.
In general, we set $k$ to be roughly proportional to $t$, controlled by a compression ratio $r \approx t/k$.
Depending on the task,
$k$ can either grow with $t$ when $w_{1:t}$ is incrementally observed~(\cref{sec:asauto}), 
or be strictly proportional to $t$ when $w_{1:t}$ is fully observed~(\cref{sec:asencoder}). 

\subsection{\model as an autoregressive LM}
\label{sec:asauto}

Not all efficient LMs support causal masking~\citep{abc21}.
Many context compression methods \citep{gist23,icae24} only apply to fixed-sized texts.
However, each hidden state $\vec z_i$ in \repr only conditions on its past tokens. Thus \model can be naturally integrated into an autoregressive LM, where tokens $w_{1:t}$ are sequentially fed into an LM.
Instead of saving all past hidden states $\vec x_{1:t}$, \model only retains a subset of tokens as \repr, which are selected by $\sco$.
The stochastic selection process in \cref{eq:zselection} is made deterministic by settings a threshold $\Lambda$ in \cref{eq:scorer}:
\begin{align}
    \alpha_i = { \mathbbm{1}} \left\{\sco_\varphi(\vec x^\iota_i) > \Lambda\right\},\label{eq:indicator}
\end{align}
where $\mathbbm{1}$ is the indicator function.
That is, token $w_i$ is retained as \repr $\vec z_j$ if its score is above the threshold $\Lambda$.
Because \cref{eq:indicator} does not depend on future tokens, 
$\vec z_{1:k}$ can be autoregressively encoded with causal masking.

To set a proper threshold $\Lambda$, we define a compression ratio $r \ge 1$ and let $r \approx t/k$.
That is, $\Lambda$ should be set such that after $t$ tokens are fed into \model, roughly $k \approx t/r$ hidden states $\vec x_i$'s should be selected as $\vec z_j$'s.
In practice, we estimate the threshold $\Lambda$ by running a trained $\sco_\varphi$ on sampled tokens.
\footnote{
Training $\sco_\varphi$ requires a determined $\Lambda$, but setting $\Lambda$ needs a trained $\sco_\varphi$.
To prevent the chicken-and-egg problem, we initialize the $\sco_\varphi$ here from \cref{sec:asencoder}.
}

\paragraph{Parameter configuration}
Intuitively, as a compressed representation, $\vec{z}_j$ should encode a broader range of tokens than $\vec x_i$ does.
We therefore separate their attention parameters:
Once a token $w_t$ is selected by \cref{eq:indicator}, it uses $\attn_\phi$ to attend past tokens.
Otherwise, it uses $\attn_\theta$.

\begin{figure}
    \centering
    \includegraphics[width=7.7cm]{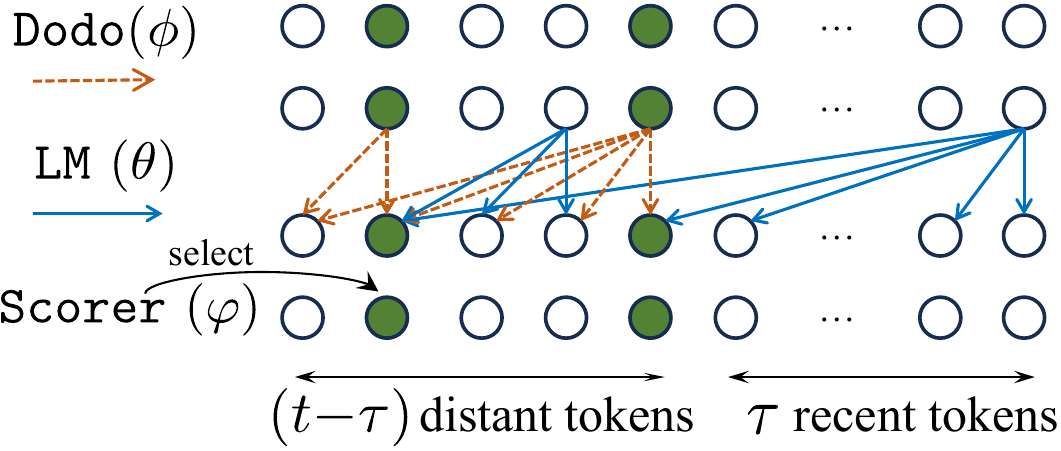}
    \caption{
    An illustration of the autoregressive \model,
    where $\sco(\varphi)$ selects \repr tokens,
    $\nugop(\phi)$ aggregates the information of $(t-\tau)$ distant tokens into \repr.
    When predicting a new token, the $\mathtt{LM}(\theta)$ has direct access to recent $\tau$ tokens but needs to use \repr to access the distant information.
    }
    \label{fig:autoregressive}
\end{figure}

\paragraph{A mixed resolution}
Though $\vec z_{1:k}$ is more efficient than $\vec x_{1:t}$, information loss is inevitable during the subselection process.
Intuitively, the tokens closer to the target token $w_{t+1}$ contain more relevant information.
We propose to revise \cref{eq:zattn} with a mixed resolution, where \emph{$\vec{x}_t$ attends to recent $\tau$ tokens without compression}.
Suppose we split the sequence $w_{1:t}$ at index $(t-\tau)$, we have
\begin{align}
    \vec x_{t}^{l+1} &= \attn_\theta\left(\vec{x}_t^l, \left[\vec z_{1:k}^l;\vec x_{t-\tau:t}^l\right]\right), \label{eq:mixedreso} \\
    \vec z_{1:k} &= \nugop_{\phi,\varphi}(w_{1:t-\tau}) \label{eq:mixedreso2}
\end{align}
where $\vec z_{1:k}$ are the compressed representation of $w_{1:t-\tau}$, $[~~;~~]$ indicates the concatenation of vector sequences,
and $\tau$ is a hyperparameter.
An illustration of our method can be seen in \cref{fig:autoregressive}.

\paragraph{Learning}
To train \model as an autoregressive LM, we estimate the parameters $(\theta,\phi,\varphi)$ to maximize the log likelihood of $p(w_{1:n})$:
\begin{align}
    \max_{\theta,\phi,\varphi} \ \sum_{w_{1:n}\in\mathcal{D}}\ \sum_{i=1}^{n-1} \log p(w_{i+1} \mid w_{1:i}),\label{eq:autoobj}
\end{align}
where $\mathcal{D}$ is the corpus and $p(w_{i+1} \mid w_{1:i})$ is defined by \cref{eq:lmhead,eq:mixedreso,eq:mixedreso2}.

Learning with \cref{eq:autoobj} can be inefficient: The computation cannot be parallelized on the sequence dimension because they have different splitting index $(i-\tau)$.
As an efficiency optimization, we chunk the texts into segments, and tokens in a segment share the same splitting index.

\subsection{\model as a contextual compressor}
\label{sec:asencoder}

\begin{figure*}[t]
\centering
\includegraphics[width=16cm]{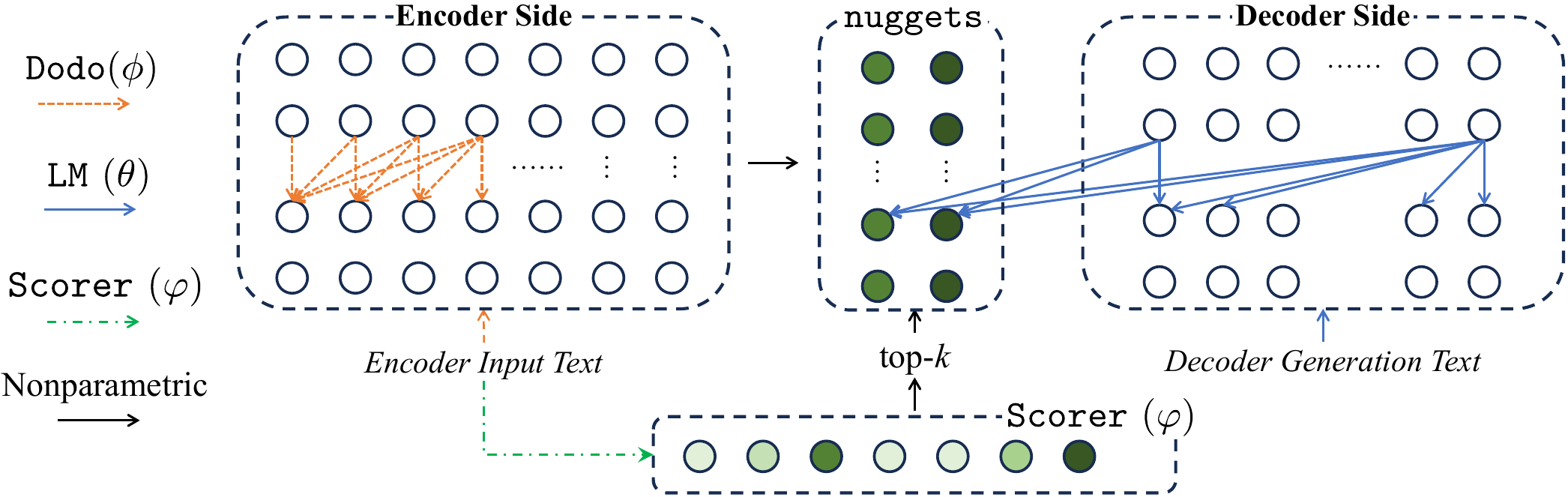}
\caption{
\model as context compressor.
From left to right,
\textbf{Encoder side}: $\nugop_\phi$ encodes texts into vectors representations;
\textbf{Scorer}: $\sco_\varphi$ computes a score for eaceh encoder token and then select the top-$k$ tokens as \repr;
\textbf{Decoder side}: Language model $\texttt{LM}_\theta$ autoretressively decodes text conditioned on \repr.
}
\label{fig:dqa}
\end{figure*}

In some tasks, such as long-form question answering, a fixed segment text, say $w_{1:n}$, is treated as the context and is fully observed before the text generation.
In this case, one can use \model as an encoder
\footnote{We use the term ``encoder'' because it encodes an input sequence. It is technically a decoder-only transformer model.}
to encode the input text into hidden states $\vec z_{1:k}$ where $k\le n$.

Formally, suppose $w_{1:n}$ and $y_{1:m}$ are the input and output sequences separately, the probability distribution of $y_{1:m}$ is defined as
\begin{align}
    &p(y_i \mid y_{<i}, w_{1:n}) \sim \lmhead_\theta\left(\vec y_i^L\right), \label{eq:ywprob}
    \\
    &\vec y_i^{l+1} = \attn_\theta\left(\vec y_i^l, \left[ \vec{z}_{1:k}^l  ; \vec y_{1:i}^l \right]\right), \label{eq:yattn}
\end{align}
where we slightly abuse the notation to use $\vec y_i$ as the hidden states of token $y_i$.
Refer to \cref{fig:dqa} for an illustration of \cref{eq:yattn}.

Because $n$, the number of input tokens, is known, we could maintain a fixed compression $r=n/k$ by setting $k=\lceil n/r \rceil$.
We therefore make the stochastic selection in \cref{eq:scorer} deterministic by:
\begin{align}
    \{\vec z_1, \dots, \vec z_k\} &= \topk(\vec x_{1:n}, s_{1:n},k), \label{eq:topk}\\
    s_i &= \sco_\varphi(\vec x_i^\iota), \label{eq:topkscorer}
\end{align}
where $\topk$ selects $k$ vectors from $\vec x_{1:n}$ with the highest $s_i$, the score of token $w_i$.
\footnote{
Because $\vec x_i$ only encodes texts before $w_i$, the last token $w_n$ is always selected to the information in $w_{1:n}$ is completely encoded in $\vec z_{1:k}$.
}

\paragraph{Parameter configuration}
We assign separate parameters to the attention modules in the encoder and decoder: 
The parameters of the encoder (decoder) are indicated by $\phi$ ($\theta$).

\paragraph{Learning} 
To train \model as an encoder, we learn it through maximum likelihood estimation:
\begin{align}
    \max_{\theta,\phi,\varphi} \sum_{w,y\in\mathcal{D}} \sum_{i=1}^{m} \log p\left(y_i\mid y_{<i},w_{1:n}\right),\nonumber
\end{align}
where input and output sequence pairs $(w_{1:n},y_{1:m})$ are sampled from a corpus $\mathcal{D}$, and the next-token probability is defined by \cref{eq:ywprob,eq:yattn,eq:topk,eq:topkscorer}.

\subsection{Learning with straight-through estimator}
\label{sec:residual}

The selection of $\vec z$ is discrete: the selection process, \cref{eq:indicator,eq:topk}, is \emph{not differentiable}.
Here we show how to back-propagate the gradients so the parameter $\varphi$ in $\sco_\varphi$ can be learned.

Previous work proposed approaches to make  $\topk$ differentiable (e.g., \citealp{softopk20} and \citealp{sander2023FastDifferentiableSparse}).
To avoid unnecessary complexity, 
we adopt the biased but simpler straight-through estimator of \citet{straightthrough13}.
Suppose the token $\vec x_j$ attends to the compressed representation $\vec z_i$, and let $\xi_{i,j}$ denote the logit of the attention token $\vec x_i$ to the compressed hidden state $\vec z_j$. Then we have (c.f. \S3.2 in \citealp{nugget23} and \S2.2 in \citealp{gumbel17}):
\begin{align}
    \xi_{i,j}^l &= \left( \vec W_\text{Q}\vec x_j^l \right)^\top \left(\vec W_\text{K}\vec z_i^l\right),\label{eq:xi} \\
    \frac{\partial \ell}{\partial s_i} &\gets \sum_{j}\sum_{l=1}^L \frac{\partial \ell}{\partial \xi_{i,j}^l}\label{eq:straightthrough},
\end{align}
where $\vec W_\text{Q}$ and $\vec W_\text{K}$ are parameters of the self-attention, and $\partial \ell / \partial s_i$ is set to be the aggregation of the gradients of $\xi_{i,j}^l$ from future tokens in all layers.
Intuitively, $\sco_\varphi$ learns to select tokens that are more attended by future tokens.
To implement \cref{eq:straightthrough}, we replace $\xi_{i,j}^l$ in \cref{eq:xi} with:
\begin{align}
    \overline{\xi}_{i,j}^l = \xi_{i,j}^l + s_i - \stopg(s_i), \label{eq:stopgrad}
\end{align}
where the $\stopg(s_i)$ detaches $s_i$ from backward pass and ensures that the addition of $s_i$ to $\xi_{i,j}^L$ does not affect the forward pass.

\section{Overall experiment setup}

We adopt the decoder-only transformer architecture of \llama~\citep{llama23,llama223} as our base model.
For the autoencoding experiment, we use the checkpoint of \texttt{LLaMA-7B} following the baseline model ICAE~\citep{icae24}.
We use the checkpoint of \texttt{LLaMA-2-7B} for the autoregressive language modeling experiments~(\cref{sec:lm_exp}) and \texttt{LLaMA-2-7B-chat}~(\cref{sec:qa}) for the downstream NLP tasks.

We adopt \textsc{LoRA}~\citep{lora22} with a rank of 32 to fine-tune the parameters of the LM, namely $\theta$ and $\phi$.
We adopt the implementation of huggingface/PEFT packakge~\citep{peft22}.
More specifically, we fix the original parameters of \llama and add two \textsc{LoRA} adapters for $\theta$ and $\phi$ respectively. 
Different adapters are activated for the computation of compressing and decoding of \model.
We disable the adapters to produce the features to $\sco$.

We employ mixed precision to save GPU memory.
The training is scaled up to 16 NVIDIA V100 cards with DeepSpeed~\citep{deepspeed20}.
See  \cref{app:training_details} for further training details, including hyperparameters, and parameter counts.

\section{Autoencoding experiment}
\label{sec:autoencoding}

\subsection{Task, dataset, and experiment setups}

In this section, we use \model as a context compressor (\cref{sec:asencoder}) and apply it to the autoencoding task.
As a comparison, we use In-Context AutoEncoder~\citep[ICAE]{icae24} as a baseline model.
In this task,
a model is asked to reconstruct the input text from a compressed representation.
Following ICAE, we fine-tune the \texttt{LLaMA-7B} model on the Pile~\citep{pile20} dataset.
We manually split the corpus into train, dev, and test splits, and train the model until convergence.

As stated in \cref{sec:asencoder}, we use \model to compress the input text into fewer hidden states $\vec{z}$, and then use the LM to decode the input sequence.
The size of hidden states $\vec{z}$, i.e. $k$, is set to be proportional to the length of the input sequence: 
$k = n/r$, and we set $r=20$ and $10$.
We prepend a trainable soft token to the decoding sequence to signal the model to reconstruct inputs~\citep{icae24}.

The key idea of ICAE is to append 128 tokens to the input sequence as ``memory slots,'' and train the decoder to reconstruct the input from the memories:
\begin{align}
    (\tilde{\vec{m}}_1, \tilde{\vec{m}}_2, \dots, \tilde{\vec{m}}_{128}) &= \mathtt{LM}\left([w_{1:n};m_{1:128}]\right) \nonumber \\
    p(w_{i+1} \mid w_{1:i}) &= \mathtt{LM}\left([w_{1:i};\tilde{\vec{m}}_{1:128}]\right). \nonumber
\end{align}
We measure using  BLEU~\citep{bleu02} score on  pairs of input and decoded texts.
\footnote{
We report ICAE results per the \S3.3.1 in \citet{icae24}.
}

\subsection{Experiment results}

\begin{figure}
    \centering
    \includegraphics[width=7.6cm]{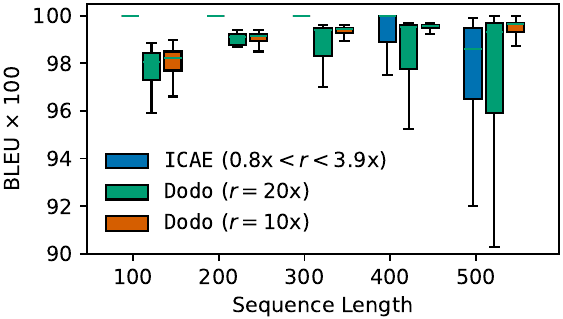}
    \caption{
   BLEU scores for autoencoding.
   Each group corresponds to a sequence length ($\pm 5$ tokens).
   Note the performance of ICAE is nearly 100\% for sequence lengths shorter than 300.
   }
\label{fig:bleu_box}
\end{figure}

In \cref{fig:bleu_box} we see \model has comparable performance with the ICAE baseline for short sequences and better performance for long sequences.
Moreover, \model successfully handles longer inputs: performance improves on longer sequences because the number of \repr is proportional to the sequence length, unlike ICAE's constant-sized memory. Despite its variable memory, \model maintains an advantage over ICAE in computational time and space.  First, \model \emph{encodes} sequences more efficiently: while ICAE always \emph{appends} 128 tokens, \model \emph{reuses} a fraction of the already-encoded tokens.  Also, \model \emph{uses fewer tokens} than ICAE: even for the longest sequences, \model only uses 25 or 50 tokens, while ICAE uses 128 for all sequences.
\footnote{
\model uses all layers while ICAE only uses the last layer. However, ICAE needs to encode their memory tokens into hidden states during decoding, while \model can save this step.
}
Lastly, \model is more efficient than ICAE during \emph{decoding} because it uses fewer tokens and does not need to re-encode them. In short, compared to the baseline, \model demonstrates comparable or better performance, successful handling of long sequences, and much more efficient encoding and decoding.

We also conducted experiments on languages other than English.
For more details, readers may refer to \cref{app:multilingual}.

\subsection{\model selects clausal text delimiters}
\label{sec:intrinsic}

\begin{figure}[t]
    \centering
    \includegraphics[width=7.6cm]{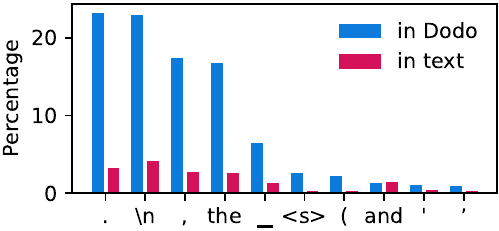}
  \caption{
   Token frequency of tokens selected by \model and the formal texts.
   These top 10 token types cover 95\% of the observed selection.
   }
\label{fig:freq_bar}
\end{figure}

In \cref{sec:n2dmethod}, we employ $\sco$ to pick out \repr,
but what are the actual tokens selected?
We empirically sampled 128 documents with 50k tokens and run the $\sco$ from the checkpoint in \cref{sec:autoencoding} with a compression ratio of $10$, and the results are shown in \cref{fig:freq_bar}.
Readers may refer to \cref{app:example_text} for case studies on sampled texts.
From \cref{fig:freq_bar}, we observe similar phenomena as \citet{nugget23}, 
where the tokens preferred by \model are mostly clausal text delimiters, 
such as punctuation marks and conjunction words.
This phenonenon is further discussed in \cref{sec:discussion_delimiters}.

\begin{figure*}
    \centering
    \fbox{
    \begin{minipage}{15.5cm}
    \small
    \textit{\color{blue}
    \dots In the 1890s, armed standoffs were avoided narrowly several times.
    \underline{The Great Northern Railway}, under the supervision of president \dots (omitted 230 tokens) \dots The railway also built Glacier Park Lodge, adjacent to the park on its east side, and the Many Glacier Hotel} \textit{\color{red}on the east shore of Swiftcurrent Lake. Louis Hill personally selected the sites for all of these buildings, choosing each for their dramatic scenic backdrops} {\color{red}  and views. Another developer, John Lewis, built the Lewis Glacier Hotel on Lake McDonald in 1913–1914.} \textbf{The Great Northern Railway} {\color{gray} bought the hotel in 1930 and it was later \dots}
    \end{minipage}
    }
    \caption{
    An example of a setting of our LM experiment. Here, compressive models access 320 tokens of  history (italics) which they must compress to 32 states, along with 32 explicit tokens of most recent history (final portion of red, normal text). \full gets explicit access only to the entirety of the red text (64 tokens), with no access to longer history. Models need to complete the sequence starting with \textbf{The Great Northern Railway}. 
    }
    \label{fig:auto_example_text}
\end{figure*}

\begin{table*}[ht!]
    \centering
    \begin{tabular}{l|c|cc|cc|c}
    \thick
       \multirow{2}{*}{model}  &  total & compressed &  context  &\multicolumn{2}{c|}{ppl. on WikiText} & ppl. on Pile \\
       &states&tokens&tokens&\textit{subword}&\textit{word}&\textit{subword} \\
       \hline
       \full & 256 & 0 & 256 & 6.39 & 10.65  & 4.94
       \\
       \compressive  & 256 & 1280 & 128 & 6.88 & 11.62 & 4.82
       \\
       \model & 256 & 1280 & 128 & \textbf{6.30} & \textbf{10.55} & \textbf{4.01} \\
       \hline
       \full & 128 & 0&128 & 6.87 & 11.69  & 5.35
       \\
       \compressive  & 128 & 640 & 64 & 7.09 & 12.18 & 4.93
       \\
       \model & 128 & 640 & 64 & \textbf{6.58} & \textbf{11.06} & \textbf{4.49}
       \\
       \hline
       \full & 64 & 0 & 64 & 7.95 & 14.08 &  5.80
       \\
       \compressive  & 64 & 320 & 32 & 7.64 & 13.39 & 5.65
       \\
       \model & 64 & 320 & 32 &\textbf{6.91}&\textbf{11.78}& \textbf{5.01}
       \\
       \thick
    \end{tabular}
    \caption{
    Perplexity on the Pile and WikiText-103, contrasting two 10x compressed solutions against no use of compression. 
    \textbf{Compressed tokens}: the number of compressed tokens that precede \textbf{context tokens}: the uncompressed context immediately before the token to be predicted. This adds up to \textbf{total state}, which is directly comparable between systems, using three settings (256, 128, and 64). \model trades off explicit context for larger history, with better perplexity results.
    }
    \label{tab:wiki}
\end{table*}

\section{Autoregressive LM experiment}
\label{sec:lm_exp}

\subsection{Experiment setup}

In this task, the model is asked to \emph{autoregressively} decode a sequence of texts.
We therefore use \model as an autoregressive LM~(\cref{sec:asauto}).
We introduce a baseline method Compressive Transformers~\citep{compressive20} (denoted by \compressive),
which evenly chunks the text into segments and uses a pooling algorithm
\footnote{In experiments, we adopt the mean pooling.}
to compress the hidden states of each segment into a single vector.
We also conduct experiments with the original \llama, denoted by \full.
In experiments,  \compressive has the save compression ratio as \model does.
\full does not support compression, so we limit its context length to make sure all models use the same number of hidden states.

We use the Pile~\citep{pile20} and WikiText-103~\citep{wikitext16} as the corpus.
We randomly split the Pile into train, dev, and test sets, where the test set contains 100k tokens.
All models are initialized from the checkpoint \texttt{Llama-2-7b}, and trained on the training set of the Pile until convergence.
The compression ratio for \model and \compressive is 10x.
The evaluation is conducted on the test set of the Pile and WikiText-103.

Perplexity (PPL) is used as the evaluation metric.
Following previous work, we exclude the words that are defined as out-of-vocabulary by \citet{wikitext16} from the evaluation on WikiText-103. 
Because WikiText-103 is a tokenized corpus, we take production over the probabilities of subwords for each complete word to measure the word PPL.
Note our algorithm underestimates the model performance for the complete word PPL.

We illustrate the intuition of \model via an example in \cref{fig:auto_example_text}. For such an example, \model  should retain both topical and explicit vocabulary information (e.g., the underlined text) in the compressed history, in order to be less surprised by subsequent text such as bolded there.

\subsection{Experiment results}

The experiment results are shown in \cref{tab:wiki}.
We conduct experiments with 3 context configurations,
where an LM has access to up to 64, 128, or 256 past hidden states.
For \model and \compressive, the first 32, 64, or 128 states are compressed representation of the past 320, 640, or 1280 tokens.
\model outperforms both \compressive and \full, showing that with a restricted size of hidden states, \model is an effective method to encode history information.

\begin{table*}[t]
    \centering
    \begin{tabular}{l|ccc|ccccc}
    \thick
    \multicolumn{1}{c|}{\multirow{2}{*}{Dataset}} & \multicolumn{3}{c|}{Split sizes} & \multicolumn{3}{c}{Text length}  \\
    & train & dev & test & doc & query & answer\\
    \hline
    SQuAD~\citep{squad16}&88k&10.5k&-&231&17.0&-\\
    CNN/DailyMail~\citep{pointgen17} &287k&13.4k&12k&878 & - & 68.9\\
    \thick
    \end{tabular}    
    \caption{
    Dataset statistics. The text lengths are counted by the LLaMA tokenizer.
    }
    \label{tab:data_stats}
\end{table*}

\section{Downstream task experiments}
\label{sec:qa}

We pick downstream tasks where a document as context is followed by a query.
The model is asked to encode the document and decode the answer conditioned on the document encoding and question.
In these tasks, we use \model as a context compressor~(\cref{sec:asencoder}),
and we set the compression $r=5$ or $10$.
To train \model to perform these tasks, we consider 2 scenarios.
a) \textbf{Fine-tuning}: \model is trained on the training set of the downstream tasks.
b) \textbf{Zero-shot}: \model is trained on normal texts randomly sampled from the Pile and directly tested on the downstream task.
In this case, each text is split into 2 parts, containing up to 512 and 128 tokens, and the model is asked to decode the second part conditioned on the encoding of the first part.

We consider the tasks of question answering and summarization.
Datasets used in this section are SQuAD~\citep{squad16} and CNN/DailyMail v3.0.0~\citep{pointgen17} for summarization.
Their statistics are listed in \cref{tab:data_stats}.

We use the following baseline methods:
\begin{itemize}
    \item \full: Results of the original LM.%
    \item \nodoc: LM is used to do the task without any documents. Only the question is provided.
    \item \summ: Use the LM to summarize the text into fewer tokens with prompts, 
    which asks the LM to compress the texts into 10\% of its length.
    LM uses the summary instead of documents to do the task. (\cref{app:lmsummprompt})
    \footnote{In practice, LM uses 10.9\% of its original length to summarize the text on average, counted by subwords.}
\end{itemize}

\subsection{Question answering}
\label{sec:squad}

\begin{table}[htbp]
\centering
\begin{tabular}{lcrr}
    \thick
     Model & cmpr. & accuracy \\
     \hline
     \nodoc & $\infty$ & 1.4 \\
     \summ & 10x & 30.9 \\
     \full & 1x & \textbf{64.5} \\
     \hline
     \model & 5x & 59.1 \\
     \model & 10x & 49.8 \\
     \thick
\end{tabular}
\caption{
The accuracy of all 4 models on the task of SQuAD.
Cmpr. is the compression ratio of the method.
}
\label{tab:squad_results}
\end{table}
In SQuAD a model is asked to extract a phrase from the passage to answer the query.
We reformulate this problem as a text-to-text task instead of annotation and prompt the model to answer the question~(\cref{app:squadprompt}).
We use accuracy to evaluate the model performance.
As the model tends to generate tokens more than the answer itself or using different forms (e.g. using ``two'' instead of ``2''), we normalize the output to match the answer. 
Readers may refer to \cref{app:squad_normalization} for the algorithm used to calculate the accuracy.

We consider all models: \full, \summ, \model, and \nodoc~(\cref{tab:squad_results}).
All models are evaluated in a zero-shot manner without fine-tuning.
\full and \model easily outperform the \nodoc and \summ,
and we observe that \summ often omits details that are needed by the question.
The performance of \model can be improved by lowering its compression ratio,
and the performance of \model ($r=5$) is close to \full, confirming a compressed representation can still support LLM reasoning.

\subsection{Summarization}
\label{sec:summ}

\begin{table}[thbp]
    \centering
    \begin{tabular}{lrrrr}
    \thick
    model & cmpr. & R1 & R2 & RL \\
    \hline
    \full (zero-shot) & 1x & 32.5 & 9.7  & 28.2 \\
    \full (fine-tuning)  & 1x &37.7 & \textbf{15.6} & 35.3 \\
    \model & 10x &\textbf{39.9} & 14.6 & \textbf{37.0} \\
    \thick
    \end{tabular}
    \caption{
    The Rouge scores (F$_1$ of Rouge-1, Rouge-2, LCS) of  \full and \model on CNN/DailyMail.
    }
    \label{tab:cnn}
\end{table}

CNN/DailyMail contains news articles, where a model is required to generate a short summary.
As no query is involved, we propose a prompt as a statement of the task requirement~(\cref{app:summprompt}).

We consider \full and \model~($r$\,=\,10).
\full is evaluated in both zero-shot and fine-tuning settings and \model is fine-tuned.
The results are shown in \cref{tab:cnn}.
We find that \model can achieve similar or even better performance than \full after compression.
We speculate that as the context of CNN/DailyMail is long, this may lead the LM to be ``lost in the middle''~\citep{lost24}, whereas the \repr generated by \model is only 10\% of the original length and perhaps less susceptible.  
This is an interesting avenue for future exploration.

\section{Discussion}

\subsection{The selection of \repr}
In \model, $\sco$ selects $k$ vectors out of $n$ candidates at each layer of the transformers.
We adopt a solution of \emph{hard selection} because of its simplicity.
Some alternatives, such as soft attention and soft top-$k$ operator, require either additional parameters or advanced machine learning techniques.
Hard selection learns to naturally split the text, which contrasts some pooling strategies that evenly split the text (c.f. \cref{sec:lm_exp}).

\nugget selection is learned through the residual connection introduced in \cref{sec:residual}.
With gradient signal from the self-attention, $\sco$ tends to select the tokens that are mostly attended by the decoder.
Isolating the other parts of the model,
\emph{how can we evaluate the performance of $\sco$ itself}?

To simplify the discussion, let $\mathcal{I}$ be the selection conducted by $\sco$.
We use \istar to denote the \emph{theoretically optimal \repr\, selection}, which is defined as the selection that achieves the best performance in a task, e.g. the lowest perplexity in the LM task.
To evaluate $\mathcal{I}$, we ask:
How similar are $\mathcal{I}$ and \istar? 
What is their performance gap?

Unfortunately, finding the optimal selection \istar is a non-trivial combinatorial problem, so we propose a greedy algorithm to approximate \istar.
Due to the space limit, we leave the details of this algorithm and our experiment design to \cref{app:optimal}.
As the results, the overlapping between $\mathcal{I}$ and \istar is roughly 75.3\%, meaning the \repr selected by $\sco$ are very close to the theoretical optimal selection.
Replacing \istar with $\mathcal{I}$ will sacrifice 7.9\% of the performance in terms of LM perplexity,
so we conclude that $\sco$, though not being optimal, can achieve a near-optimal performance through the straight-through estimator.

\subsection{\model favors clausal text delimiters}
\label{sec:discussion_delimiters}

In \cref{sec:intrinsic}, we observed that \model favors clausal text delimiters as the \repr tokens,  similar to the findings of \citet{nugget23}.
We have the following assumptions:
\begin{itemize}
    \item \emph{Clausal text delimiters are used as ``summarization tokens'' during pretraining}. The LM was pretrained to predict the next token, and predicting the text delimiters was equivalent to predicting the ending of a clause/sentence. Therefore, the LM learned to store contextual information in the delimiters, such as punctuation marks.
    \item \emph{$\sco$ was biased to frequent tokens.} Except for the clausal text delimiters, \model also prefers the token ``the'', which hints that the straight-through estimator in \cref{sec:residual} might bias $\sco$ to select frequently appeared tokens.
\end{itemize}

\section{Related work}
\label{sec:related}

\subsection{\nugget text representation}

\model can be viewed as a natural extension of \nugget on \emph{decoder-only} transformers.
They are similar regarding the vector subselection~(\cref{sec:n2dmethod}) but different in architecture and applications.
From the perspective of \emph{architecture}, different from \nugget that reduces the last-layer representation of a transformer encoder, \model reduces the memory and computation of self-attention in a transformer decoder.
Also, \model replaces the residual connection used by \nugget with straight-through estimator~(\cref{sec:residual}), which naturally cancels the side-effect of the residual connection in the forward pass.
From the perspective of \emph{applications}, 
because \model supports causal masking, it can be used for autoregressive language modeling without re-computation.
\nugget, instead, is more suitable for text similarity measurement.

\subsection{Scaling the context length of transformers}

Scaling transformers to long sequences is a popular topic in the NLP community~\citep{efficient22}. 
Existing work includes sparsify the attention patterns~\citep{longformer20,zaheer2020BigBirdTransformers,chordmixer23,ding2023LongNetScalingTransformers,ainslie2023CoLT5FasterLongRangea,compressive20}, employing low-rank or kernel methods to approximate the attention matrix computation~\citep{performer21,katharopoulos2020TransformersAreRNNs},
or applying recurrence~\citep{transformerxl19,xlnet19,rmt22}.
Another line of work tries to extrapolate the ability of LMs to long contexts, such as using linear bias~\citep{press2022TrainShortTest} or rotary position embeddings~\citep{roformer24}.
Recently, \citet{unlimiformer23,longllama23} applied $k$NN search to select a subset of tokens for attention at each layer of an encoder-decoder transformer, effectively extending the attention range of transformers.
\citet{vcc23} proposed to compress the context by prioritizing the ``VIP tokens'', which are important to certain tasks and can be saved in specialized data structure.

Past work on efficient transformers, as shown above, mainly improves the efficiency of the self-attention.
\model instead addresses a language representation problem: It shortens the length of the sequences in the space of hidden states.
From this perspective, the idea of \model is orthogonal to most of the efficient self-attention methods, and thus can be jointly applied with most of them, e.g. $k$NN based methods~\citep{longllama23}.

In the context of large language models, recent work focuses on compressing the prompt tokens into soft embeddings~\citep{gist23,wingate2022PromptCompressionContrastive} or encoding the supporting documents~\citep{icae24,chevalier2023AdaptingLanguageModels} into fewer vectors.
LLMLingua~\citep{llmlingua23} is a coarse-to-fine prompt compression method that allocates different compression ratios over various prompt components.
Some recent work tries to train LLMs with longer contexts, such as \citet{longchat23}, GLM~\citep{glm23}, and Claude 2~\citep{claude23}.
Notably, \citet{xiong2023EffectiveLongContextScaling} continue to train \textsc{LLaMA} to study the relationship between model performance and context length.

Researchers also explored retrieval-based methods that infuse knowledge into LM decoding,
some notable work in this field includes FiD~\citep{fid21}, REALM~\citep{realm20}, KNN-LM~\citep{knnlm20}, and RAG~\citep{rag20}.
From the angle of the LLMs, \citet{zheng2023WhyDoesChatGPT} found that providing contexts to LLMs can help them generate truthful answers. %

\section{Conclusion}

In this work, we propose \model, a method for contextual compression for decoder-only transformers.
In language modeling~(\cref{sec:lm_exp}) and summarization~(\cref{sec:summ}),
\model is shown to generate a highly condensed representation of the context,
while the results in autoencoding~(\cref{sec:autoencoding}) and question answering~(\cref{sec:squad}) reflect that the details of the contexts can be recovered from \repr.
Moreover, in \cref{sec:squad} we show that \model trained with text continuation preserves the capability of instruction following.
This demonstrates LLMs can encapsulate more of their input into fewer hidden states than previously realized, suggesting a new direction for efficient foundation models.
Future work will explore more specialized versions of this proposal for optimizing results on individual applications, such as in dialog, supervised fine-tuning, reinforcement learning with human feedback, and in-context learning.

\section*{Ethical statement and limitations}

\paragraph{Used artifacts}
In this work, we used the publicly released codes and checkpoints of \llama.
Per the license attached to \llama, we agree not to re-distribute their parameters and limit the usage of the models for research purposes only.

\paragraph{Potential societal risks}
Because we only trained \llama\, on general texts, we do not think that our paper will have any additional societal impacts beyond the checkpoints, except for the privacy issues mentioned below.

\paragraph{Privacy issues on the datasets}
Our method further fine-tunes \llama\, on the Pile~\citep{pile20}.
Given the size of the Pile~\citep{pile20} is huge (around 800GB), we are unable to conduct effective investigations on the privacy issue on the corpus.
We refer readers to \citet{pile20} for the discussion of the potential issues inside the data.

\section*{Acknowledgment}

We thank Ho-Lam Chung and Canwen Xu for their thoughtful discussion.
We thank William Fleshman for his valuable feedback on the writing.

This work has been supported by the U.S. National Science Foundation under grant no. 2204926. Any opinions, findings, conclusions, or recommendations expressed in this article are those of the authors and do not necessarily reflect the views of the National Science Foundation.

\bibliography{ref}
\bibliographystyle{acl_natbib}

\clearpage
\newpage
\appendix

\section{Optimal \repr selection}
\label{app:optimal}

The \repr selection module, i.e. $\sco$, is learned through the residual connection introduced in \cref{sec:residual}.
With gradient signal from the self-attention, $\sco$ tends to select the tokens that are mostly attended by the decoder (parameterized by $\theta$).
However, it remains a question whether the selection is optimal.
Here we provide an empirical estimate of the gap between the optimal \repr selection and $\sco$.

Suppose we select $k$ \repr out of $n$ tokens, we define a selection as a set of indices 
$$\mathcal{I}=\{i_1, i_2, \dots, i_k\},\quad 1 \le i_j \le n.$$
From the definition, we can see that
$$\mathcal{I} \subseteq \{1,2,3,\dots,n\}.$$
We further define the optimal selection \istar as the selection that achieves \emph{the best performance} on a downstream task, e.g. lowest perplexity for language modeling.
We denote the selection of $\sco$ as \ibar.
We want to answer two questions: How similar are \istar and \ibar, and what is the performance gap between \istar and \ibar?

Finding \istar is a non-trivial combinatorial optimization problem.
The only possible solution, as we know, is to enumerate $\binom{n}{k}$ different selections,
which is infeasible for large $n$ and $k$.
Therefore, we approximate \istar with a greedy algorithm.
The basic idea is to start with $\mathcal{I}\gets\bar{\mathcal{I}}$.
Iteratively, for each index $i\in \mathcal{I}$, we replace it with an optimal index from the un-chosen indices so that it achieves the best downstream performance.
We formalize it in \cref{alg:optimal} with an example downstream task of language modeling.

\renewcommand{\algorithmicrequire}{\textbf{Input:}}
\renewcommand{\algorithmicensure}{\textbf{Output:}}
\begin{algorithm}
\caption{A greedy algorithm to find the ``optimal'' selection \istar.}\label{alg:optimal}
\begin{algorithmic}
\Require $k$ (number of \repr) and $n$ (number of tokens) ($0 < k \le n$), encoder outputs $\vec{x}_{1:n}$
\Ensure A selection $\mathcal{I}$ and the corresponding LM perplexity $b$
\State Initialize $\mathcal{I}=\{i_1, i_2, \dots, i_k\}$ with $\sco$.
\State Perplexity $b\gets \mathtt{Decoder}(\vec{x}_{1:n},\mathcal{I})$
\Comment{Lowest perplexity so far}
\For{$i \in \mathcal{I}$} 
\For{$i'\in \{1,2,\dots,n\}\backslash \mathcal{I}$}
\Comment{All possible replacements from unchosen indices}
\State $\mathcal{I}'\gets \left(\mathcal{I}\backslash\{i\}\right)\cup\{i'\}$
\Comment{Replace $i$ in $\mathcal{I}$ with $i'$}
\State Perplexity $b'\gets \mathtt{Decoder}(\vec{x}_{1:n},\mathcal{I}')$
\If{$b' < b$}
\Comment{If $i'$ is better than $i$, make the replacement permanent}
\State $b\gets b'$, $\mathcal{I} \gets \mathcal{I}'$
\EndIf
\EndFor{}
\EndFor
\end{algorithmic}
\end{algorithm}

We conduct experiments with the checkpoints in \cref{sec:lm_exp}.
We compress a sequence of up to 128 tokens into \repr with a compression ratio of 10x.
We present the model with another 64 tokens without compression.
The model is required to predict the next 64 tokens, and we measure the subword-level perplexity of \model.
Because \cref{alg:optimal} contains 2 for loops and is expensive to execute, we only sample 1000 documents from the test set of WikiText-103~\citep{wikitext16}.

To measure the difference between \ibar and \istar, we count how many elements are replaced in \ibar with \cref{alg:optimal}.
On average, 24.7\% \repr tokens are replaced, meaning $\sco$ is roughly 75.3\% ``correct''.
After replacing \ibar with \istar, the overall subword-level perplexity is improved from 7.74 to 7.13, or \istar is roughly 7.9\% better than \ibar in terms of downstream task performance.

In conclusion, we conduct experiments to show that $\sco$ is adequate to select \repr as it can achieve similar performance as a decoder-aware optimal selector.

\section{Implementation \& training details}
\label{app:training_details}

\subsection{Implementation}

The training pipeline of \model is implemented with the PyTorch~\citep{pytorch19} and Pytorch Lightning package~\citep{lightning19}.
We use the ZeRO stage-2 provided by the DeepSpeed~\citet{deepspeed20} package with mixed precision to accelerate the training.
The implementation of \model is based on the huggingface/transformers package~\citep{hftransformers20}.
Our dataset reader uses huggingface/datasets~\citep{datasets21}.

\subsection{Hyperparameters and training devices}

For all the experiments, we follow the training setup of \citet{llama223} and use an Adam optimizer~\citep{adam15} with a learning rate of $1\times 10^{-4}$, $\beta_1=0.9$, $\beta_2=0.95$, and $\epsilon=10^{-5}$.
We use a cosine learning rate scheduler~\citep{cosine17} with warmup of 2k steps, and the period of the cosine annealing function is set as 150k steps.

All the text generation processes in this paper are implemented as greedy decoding.

We train the models on 16 NVIDIA Tesla V100 GPUs (32 GiB), each with a batch size of 1.
Gradients are accumulated for 2 batches before the execution of the optimizers.
All the models are trained until early stopping because of the convergence of the loss on the validation set.

\begin{table}[t]
\small
    \centering
    \begin{tabular}{l|rrrr}
         \thick
         module & \#params & percentage & trainable  \\
         \hline
         \llama-7B & 6.74B & 99.01\%& no \\
         encoder ($\phi$) & 25.2M & 0.37\% & yes \\
         decoder ($\theta$) & 25.2M & 0.37\% & yes \\
         $\mathtt{Scorer}$ ($\varphi$) & 16.8M &  0.25\% & yes\\
         soft prompt ($\theta$) & 4,096 & $<$0.0001\% & yes \\ 
         \thick
    \end{tabular}
    \caption{
    Parameter count of \model. We do not distinguish \texttt{Llama-7b}, \texttt{Llama-2-7b}, and \texttt{Llama-2-7b-chat} here as they have the same architecture.
    The parameters of the encoder and decoder are counted as additional parameters with LoRA compared to the base model.
    }
    \label{tab:param_count}
\end{table}

\subsection{Number of parameters}
\label{app:param}

In this section, we enumerate the number of parameters in \model, as shown in \cref{tab:param_count}.
Except for the frozen \llama model, \model has an encoder and decoder, which contains additional parameters to the Llama model with LoRA~\citep{lora22} (rank $=$ 32), a scorer (2-layer feedforward neural networks), and a soft prompt that adds a special token to the embedding matrix.

For the experiments in \cref{sec:lm_exp}, we use LoRA to train \compressive,
which contains a decoder and a soft prompt as we have shown in \cref{tab:param_count}. 
However, compared to the size of \llama, the trainable parameters of both \model and \compressive are significantly fewer ($<$1\%).

\section{Example text for \repr selection analysis}
\label{app:example_text}
We sample a passage from Wikipedia and run $\sco$ on the text, where we set the compression ratio $r=10$.
The results are shown in \cref{fig:nugget_case}.

\definecolor{grey0}{RGB}{195,195,195}
\definecolor{grey1}{RGB}{182,182,182}
\definecolor{grey2}{RGB}{169,169,169}
\definecolor{grey3}{RGB}{156,156,156}
\definecolor{grey4}{RGB}{143,143,143}
\definecolor{grey5}{RGB}{130,130,130}
\definecolor{grey6}{RGB}{117,117,117}
\definecolor{grey7}{RGB}{104,104,104}
\definecolor{grey8}{RGB}{91,91,91}
\definecolor{grey9}{RGB}{78,78,78}
\definecolor{grey10}{RGB}{65,65,65}
\definecolor{grey11}{RGB}{52,52,52}
\definecolor{grey12}{RGB}{39,39,39}
\definecolor{grey13}{RGB}{26,26,26}
\definecolor{grey14}{RGB}{13,13,13}
\definecolor{grey15}{RGB}{0,0,0}
\definecolor{bgc}{RGB}{153,255,153}

\begin{figure*}[t]
\centering
\fbox{
\begin{minipage}{16cm}
{\color{grey14}The} {\color{grey9}Brook}{\color{grey10}lyn} {\color{grey12}N}{\color{grey9}ets} {\color{grey13}have} {\color{grey11}built} {\color{grey3}themselves} {\color{grey8}up} {\color{grey13}from} {\color{grey11}next} {\color{grey9}to} {\color{grey6}nothing}\colorbox{bgc}{{\color{grey15}.}} {\color{grey12}De}{\color{grey2}void} {\color{grey13}of} {\color{grey3}anything} {\color{grey3}close} {\color{grey11}to} {\color{grey14}an} {\color{grey5}asset} {\color{grey7}before} \colorbox{bgc}{{\color{grey15}}}{\color{grey13}2}{\color{grey13}0}{\color{grey11}1}{\color{grey10}5}\colorbox{bgc}{{\color{grey15},}} {\color{grey14}the} {\color{grey12}N}{\color{grey9}ets} {\color{grey10}had} {\color{grey11}to} {\color{grey10}make} {\color{grey6}something} {\color{grey8}out} {\color{grey11}of} {\color{grey2}nothing}\colorbox{bgc}{{\color{grey15}.}} {\color{grey11}They} {\color{grey9}have} {\color{grey7}done} {\color{grey8}so} {\color{grey1}indeed}\colorbox{bgc}{{\color{grey15},}} {\color{grey6}loading} \colorbox{bgc}{{\color{grey15}the}} {\color{grey9}ro}{\color{grey3}ster} {\color{grey13}and} {\color{grey4}asset} {\color{grey4}cup}{\color{grey8}boards} {\color{grey5}simultaneously}\colorbox{bgc}{{\color{grey15}.}} \colorbox{bgc}{{\color{grey15}}} {\color{grey1}Unfortunately}\colorbox{bgc}{{\color{grey15},}} {\color{grey8}just} {\color{grey7}as} {\color{grey4}quickly} {\color{grey7}as} {\color{grey9}Mark}{\color{grey5}s} {\color{grey8}acquired} {\color{grey2}young}{\color{grey1}sters}\colorbox{bgc}{{\color{grey15},}} {\color{grey11}he} {\color{grey1}must} {\color{grey7}also} {\color{grey7}decide} {\color{grey11}which} {\color{grey4}ones} {\color{grey2}should} {\color{grey3}stick} {\color{grey1}around}\colorbox{bgc}{{\color{grey15}.}} {\color{grey12}It}\colorbox{bgc}{{\color{grey15}’}}{\color{grey9}s} {\color{grey13}an} {\color{grey12}ar}{\color{grey9}du}{\color{grey3}ous} {\color{grey2}exercise}\colorbox{bgc}{{\color{grey15},}} {\color{grey13}and} {\color{grey10}even} {\color{grey13}t}{\color{grey2}ough}{\color{grey8}er} {\color{grey13}for} {\color{grey15}a} {\color{grey8}team} {\color{grey5}far} {\color{grey5}from} {\color{grey8}cont}{\color{grey7}ention}\colorbox{bgc}{{\color{grey15}.}} \colorbox{bgc}{{\color{grey15}}} {\color{grey8}Most} {\color{grey4}teams} {\color{grey7}reach} {\color{grey12}this} {\color{grey6}stage} {\color{grey6}just} {\color{grey6}as} {\color{grey12}they} {\color{grey8}are} {\color{grey2}close} {\color{grey10}to} {\color{grey5}play}{\color{grey1}off}{\color{grey14}-}{\color{grey6}cal}{\color{grey11}iber}\colorbox{bgc}{{\color{grey15}.}} {\color{grey14}The} {\color{grey13}N}{\color{grey9}ets} {\color{grey6}do} {\color{grey8}not} {\color{grey8}have} {\color{grey12}this} {\color{grey3}lux}{\color{grey2}ury}\colorbox{bgc}{{\color{grey15},}} {\color{grey12}and} {\color{grey1}must} {\color{grey4}evaluate} {\color{grey12}with} \colorbox{bgc}{{\color{grey15}a}} {\color{grey7}much} {\color{grey6}longer} {\color{grey7}view} {\color{grey9}than} \colorbox{bgc}{{\color{grey15}the}} {\color{grey4}average} {\color{grey1}young} {\color{grey4}team}\colorbox{bgc}{{\color{grey15}.}} {\color{grey4}Put} {\color{grey5}simply}\colorbox{bgc}{{\color{grey15},}} {\color{grey9}they} {\color{grey1}must} {\color{grey6}think} {\color{grey6}like} {\color{grey15}a} {\color{grey6}cont}{\color{grey6}ender} {\color{grey4}before} {\color{grey4}becoming} {\color{grey9}one}\colorbox{bgc}{{\color{grey15}.}} {\color{grey15}} {\color{grey13}L}{\color{grey7}uck}{\color{grey2}ily}\colorbox{bgc}{{\color{grey15},}} {\color{grey14}the} {\color{grey6}current} {\color{grey9}ro}{\color{grey4}ster} {\color{grey12}has} {\color{grey5}distinct} {\color{grey13}t}{\color{grey3}iers} {\color{grey13}of} {\color{grey1}young} {\color{grey6}players} {\color{grey12}in} {\color{grey1}terms} {\color{grey10}of} {\color{grey10}their} {\color{grey9}long}{\color{grey13}-}{\color{grey2}term} {\color{grey1}potential}\colorbox{bgc}{{\color{grey15}.}} {\color{grey14}E}{\color{grey10}ight} {\color{grey10}of} {\color{grey14}the} {\color{grey4}nine} {\color{grey5}under}{\color{grey12}-}{\color{grey9}2}{\color{grey9}5} {\color{grey5}players} {\color{grey7}can} {\color{grey11}be} {\color{grey6}split} {\color{grey10}into} {\color{grey11}two} {\color{grey12}t}{\color{grey3}iers}\colorbox{bgc}{{\color{grey15}.}} {\color{grey15}} {\color{grey9}Lock}{\color{grey11}s} {\color{grey15}} {\color{grey13}The} {\color{grey5}group} {\color{grey9}of} {\color{grey2}definite} {\color{grey3}keep}{\color{grey2}ers} {\color{grey12}is} {\color{grey2}relatively} {\color{grey2}simple}\colorbox{bgc}{{\color{grey15}.}} {\color{grey10}These} {\color{grey7}players} {\color{grey10}have} {\color{grey14}the} {\color{grey6}most} {\color{grey1}potential} {\color{grey8}of} {\color{grey14}the} {\color{grey3}current} {\color{grey12}N}{\color{grey7}ets}\colorbox{bgc}{{\color{grey15}.}} {\color{grey14}} {\color{grey4}Although} {\color{grey13}D}{\color{grey14}’}{\color{grey7}Ang}{\color{grey3}elo} {\color{grey7}Russell} {\color{grey13}has} {\color{grey10}gone} {\color{grey5}through} {\color{grey11}some} {\color{grey4}rough} {\color{grey1}patch}{\color{grey10}es}\colorbox{bgc}{{\color{grey15},}} {\color{grey11}he} {\color{grey7}has} {\color{grey5}displayed} {\color{grey6}enough} {\color{grey5}prom}{\color{grey3}ising} {\color{grey2}signs} {\color{grey12}to} {\color{grey6}war}{\color{grey3}rant} \colorbox{bgc}{{\color{grey15}the}} {\color{grey14}“}{\color{grey0}keeper}{\color{grey9}”} {\color{grey1}status}\colorbox{bgc}{{\color{grey15}.}} {\color{grey10}His} {\color{grey6}cra}{\color{grey1}fty} {\color{grey7}ball}{\color{grey13}-}{\color{grey2}hand}{\color{grey9}ling}\colorbox{bgc}{{\color{grey15},}} {\color{grey9}scoring} {\color{grey8}off} {\color{grey15}the} {\color{grey13}d}{\color{grey7}rib}{\color{grey6}ble}{\color{grey14},} {\color{grey6}shooting} {\color{grey7}off} {\color{grey14}the} {\color{grey4}catch}{\color{grey14},} {\color{grey11}and} {\color{grey3}great} {\color{grey7}passing} {\color{grey3}vision} {\color{grey5}all} {\color{grey2}make} {\color{grey8}him} {\color{grey13}an} {\color{grey1}ideal} {\color{grey3}fit} {\color{grey11}for} {\color{grey6}Ken}{\color{grey11}ny} {\color{grey11}At}{\color{grey5}kin}{\color{grey10}son}\colorbox{bgc}{{\color{grey15}’}}{\color{grey14}s} {\color{grey3}attack}\colorbox{bgc}{{\color{grey15}.}} {\color{grey15}} {\color{grey1}Being} {\color{grey14}the} {\color{grey8}No}{\color{grey12}.} {\color{grey14}}{\color{grey9}2} {\color{grey1}overall} {\color{grey3}selection} {\color{grey11}in} {\color{grey14}a} {\color{grey3}draft} {\color{grey11}is} {\color{grey4}typically} {\color{grey3}enough} {\color{grey5}cred}{\color{grey4}ibility} {\color{grey14}to} {\color{grey4}keep} {\color{grey14}a} {\color{grey8}player} {\color{grey1}around}\colorbox{bgc}{{\color{grey15},}} {\color{grey11}but} {\color{grey6}Russell} {\color{grey11}has} {\color{grey8}shown} {\color{grey1}legit}{\color{grey8}imate} {\color{grey6}flash}{\color{grey10}es} {\color{grey12}of} {\color{grey2}star} {\color{grey1}potential} {\color{grey10}as} {\color{grey5}well}\colorbox{bgc}{{\color{grey15}.}} {\color{grey12}G}{\color{grey7}iving} {\color{grey6}up} {\color{grey11}on} {\color{grey8}him} {\color{grey4}now} {\color{grey3}would} {\color{grey8}be} {\color{grey14}a} {\color{grey3}fatal} {\color{grey1}mistake}{\color{grey15}.} {\color{grey15}} 
{\color{grey4}Jar}{\color{grey5}rett} {\color{grey10}Allen}{\color{grey15},} {\color{grey12}a} {\color{grey8}ro}{\color{grey11}ok}{\color{grey4}ie} {\color{grey4}center} {\color{grey8}from} {\color{grey15}the} {\color{grey4}University} {\color{grey7}of} {\color{grey7}Texas}{\color{grey15},} {\color{grey9}has} {\color{grey5}done} {\color{grey14}a} {\color{grey1}wonderful} {\color{grey3}job} {\color{grey11}in} {\color{grey10}his} {\color{grey3}special}{\color{grey2}ized} {\color{grey1}role}\colorbox{bgc}{{\color{grey15}.}} {\color{grey6}With} {\color{grey7}super}{\color{grey4}b} {\color{grey5}athlet}{\color{grey6}ic}{\color{grey2}ism} {\color{grey11}that} {\color{grey2}allows} {\color{grey7}him} {\color{grey12}to} {\color{grey3}protect} \colorbox{bgc}{{\color{grey15}the}} {\color{grey3}rim} {\color{grey13}and} {\color{grey4}switch} {\color{grey1}onto} {\color{grey4}per}{\color{grey9}imeter} {\color{grey1}attack}{\color{grey6}ers}\colorbox{bgc}{{\color{grey15},}} {\color{grey5}Allen} {\color{grey10}is} {\color{grey1}quite} {\color{grey1}capable} {\color{grey13}of} {\color{grey3}captain}{\color{grey8}ing} {\color{grey14}a} {\color{grey1}modern} {\color{grey2}defense}\colorbox{bgc}{{\color{grey15}.}} {\color{grey15}} 
{\color{grey8}This} {\color{grey5}athlet}{\color{grey5}ic}{\color{grey3}ism} {\color{grey4}helps} {\color{grey3}him} {\color{grey9}on} {\color{grey9}off}{\color{grey8}ense} {\color{grey7}as} {\color{grey2}well}\colorbox{bgc}{{\color{grey15},}} {\color{grey5}as} {\color{grey9}he} {\color{grey2}gets} {\color{grey8}plenty} {\color{grey10}of} {\color{grey8}lo}{\color{grey5}bs} {\color{grey10}to} {\color{grey2}finish} {\color{grey8}pick}{\color{grey13}-}{\color{grey7}and}{\color{grey13}-}{\color{grey5}roll} {\color{grey2}plays}\colorbox{bgc}{{\color{grey15}.}} {\color{grey10}When} {\color{grey8}in} {\color{grey2}doubt}{\color{grey14},} {\color{grey14}the} {\color{grey11}gu}{\color{grey5}ards} {\color{grey5}can} {\color{grey12}ch}{\color{grey6}uck} {\color{grey2}it} {\color{grey4}up} {\color{grey9}to} {\color{grey7}him} {\color{grey11}for} {\color{grey14}an} {\color{grey4}easy} {\color{grey10}de}{\color{grey2}uce}\colorbox{bgc}{{\color{grey15}.}} {\color{grey14}The} {\color{grey3}vertical} {\color{grey4}dimension} {\color{grey12}of} {\color{grey5}basketball} {\color{grey10}is} {\color{grey2}rarely} {\color{grey1}appreciated} \colorbox{bgc}{{\color{grey15}.}}
    \end{minipage}
    }
    \caption{
    Example texts processed by the $\sco$ of \model.
    Darker texts have a higher score than light texts.
    The tokens in green background are selected as \repr.
    }
    \label{fig:nugget_case}
\end{figure*}

\section{Prompts used in the paper}
\label{app:prompts}

Here we list all the prompts used in \cref{sec:qa}.

\subsection{Compress texts with LMs}
\label{app:lmsummprompt}
The prompt used by the \summ method to generate a summary for a given text is:
\begin{verbatim}
[INST]
Please summarize the following
text into $WORD words: $TEXT 
[/INST]
\end{verbatim}
We replace \texttt{\$WORD} with $\lceil n \cdot r \rceil$, where $n$ is the number of words (counted by spaces) and $r$ is a desired ratio (in \cref{sec:qa}, $r$ is 10).

\subsection{Question answering on SQuAD}
\label{app:squadprompt}
In the SQuAD experiment~(\cref{sec:squad}), a prompt is used to answer a question given a document:
\begin{verbatim}
[INST]
$DOCUMENT
Based on the provided document, 
answer the following question: 
$QUESTION 
[/INST]
\end{verbatim}
We replace \texttt{\$DOCUMENT} with the context document and \texttt{\$QUESTION} with the question.

\subsection{Summarization}
\label{app:summprompt}
In the summarization experiment~(\cref{sec:summ}), we use the following prompt:
\begin{verbatim}
[INST]
$DOCUMENT
Please summarize the above 
document in one sentence.
[/INST]
\end{verbatim}
We replace \texttt{\$DOCUMENT} with the document to be summarized.

\section{Normalization algorithm for SQuAD answers}
\label{app:squad_normalization}

The output of the language model tends to have tokens other than the answer or have different forms.
For each pair of model output and SQuAD answer, we apply the following rules:
\begin{itemize}
    \item Convert all English numbers to digits. E.g. convert ``two'' to ``2''.
    \item Replace all punctuation marks with spaces.
    \item Remove side spaces on both sides.
    \item Lowercase the string.
\end{itemize}

After these steps, a program is used to check if the model output contains the answer.
We restrict the model to generate up to 64 tokens in case they generate many tokens to hit the answer.
\footnote{They rarely do, as they are not optimized to cheat SQuAD.}

\section{Multilingual autoencoding experiments}
\label{app:multilingual}

\begin{table*}[t]
    \centering
    \begin{tabular}{l|llllllll}
        \thick
         Language & English & Bulgarian & German & French & Italian & Dutch & Polish & Russian \\
         Average Length & 348 & 346 & 393 & 346 & 295 & 228 & 325 & 407 \\
         BLEU & 99.1 & 97.7 & 98.8 & 99.0 & 98.3 & 97.9 & 98.3 & 98.9 \\
         Perplexity & 1.004 & 1.040 & 1.017 & 1.011 & 1.014 & 1.021 & 1.032 & 1.032\\
         \thick
    \end{tabular}
    \caption{The results of the multilingual autoencoding experiment.}
    \label{tab:multilingual}
\end{table*}

For the autoencoding experiment, we adopt the architecture of \llama and the checkpoint of \texttt{LLaMA-7B}~\citep{llama23} and fine-tune the model on the Pile dataset~\citep{pile20}.
Both pretraining and fine-tuning corpus are heavily biased towards English, but the tremendous size of \llama enables it to process languages other than English.
In this section, we conduct experiments to test the multilingual capability of \model.

We adopt the checkpoint of \model in \cref{sec:autoencoding} with a 10x compression ratio without further fine-tuning.
We sampled 8 languages: Bulgarian, German, English, French, Italian, Dutch, Polish, and Russian.
\footnote{We did not consider non-Indo-European languages, such as Chinese and Japanese, because we found that many characters are out-of-vocabulary for \llama.}
For each language, we sampled 100 documents from the RedPajama corpus~\citep{redpajama23}.
We truncate the document if it is longer than 512 tokens.
We use BLEU~\citep{bleu02} and perplexity as our metrics.

The results are shown in \cref{tab:multilingual}.
We can observe that \model can still process other languages, even if it was fine-tuned on an English-only corpus.

\end{document}